%% file: main.tex
\documentclass[10pt,twocolumn,letterpaper]{article}
\usepackage{cvpr}              %

\input{preamble}

\definecolor{cvprblue}{rgb}{0.21,0.49,0.74}
\definecolor{mycolor}{rgb}{0, 0.4, 0.7}
\usepackage[pagebackref,breaklinks,colorlinks,citecolor=cvprblue]{hyperref}
\usepackage{pifont}%
\usepackage{multirow}
\usepackage{makecell}

\usepackage{xcolor}

\usepackage{pgfplots}
\usepackage{pgfplotstable}
\usepackage{pgf-pie}
\pgfplotsset{compat=1.8}
\usetikzlibrary{pgfplots.groupplots,patterns,decorations.pathreplacing}
\definecolor{mycitecolor}{rgb}{0, 0.4, 0.7}
\definecolor{chromeyellow}{rgb}{1.0, 0.65, 0.0}
\definecolor{darkcerulean}{rgb}{0.03, 0.27, 0.49}
\definecolor{darkorange}{rgb}{1.0, 0.55, 0.0}
\definecolor{darkmidnightblue}{rgb}{0.0, 0.2, 0.4}
\definecolor{internationalorange}{rgb}{1.0, 0.31, 0.0}
\definecolor{internationalkleinblue}{rgb}{0.0, 0.18, 0.65}
\definecolor{lightsalmon}{rgb}{1.0, 0.63, 0.48}
\definecolor{mangotango}{rgb}{1.0, 0.51, 0.26}
\definecolor{mayablue}{rgb}{0.45, 0.76, 0.98}
\definecolor{majorelleblue}{rgb}{0.38, 0.31, 0.86}
\definecolor{mediumelectricblue}{rgb}{0.01, 0.31, 0.59}
\definecolor{bananamania}{rgb}{0.98, 0.91, 0.71}
\definecolor{mossgreen}{rgb}{0.68, 0.87, 0.68}
\definecolor{mossgreen2}{rgb}{0.58, 0.80, 0.48}
\definecolor{mintgreen}{rgb}{0.6, 1.0, 0.6}
\definecolor{palegreen}{rgb}{0.6, 0.98, 0.6}
\definecolor{bananayellow}{rgb}{1.0, 0.88, 0.21}
\definecolor{bluebell}{rgb}{0.64, 0.64, 0.82}
\definecolor{carnationpink}{rgb}{1.0, 0.65, 0.79}
\definecolor{babyblue}{rgb}{0.54, 0.81, 0.94}
\definecolor{lightgray}{rgb}{0.83, 0.83, 0.83}

\definecolor{mycolor}{rgb}{0, 0.4, 0.7}

\newlength\savewidth\newcommand\shline{\noalign{\global\savewidth\arrayrulewidth
  \global\arrayrulewidth 1pt}\hline\noalign{\global\arrayrulewidth\savewidth}}

\newcommand{\cmark}{\ding{51}}%
\newcommand{\xmark}{\ding{55}}%

\def\paperID{4704} %
\def\confName{CVPR}
\def\confYear{2024}

\makeatletter
\def\@fnsymbol#1{\ensuremath{\ifcase#1\or \dagger\or \ddagger\or
   \mathsection\or \mathparagraph\or \|\or **\or \dagger\dagger
   \or \ddagger\ddagger \else\@ctrerr\fi}}
\makeatother

\title{MS-DETR: Efficient DETR Training 
with Mixed Supervision}

\author{
Chuyang Zhao$^{12}$, Yifan Sun$^1$, Wenhao Wang$^3$, Qiang Chen$^1$, Errui Ding$^1$, Yi Yang$^4$, Jingdong Wang$^{1}\thanks{Corresponding author.}$\\
$^1$ Baidu VIS \quad $^2$ Beihang University \quad $^3$ University of Technology Sydney \quad $^4$ Zhejiang University \\
\texttt{\{zhaochuyang,sunyifan01,chenqiang13\}}@baidu.com \\
\texttt{wangwenhao0716}@gmail.com, \texttt{yangyics}@zju.edu.cn \\
\texttt{\{dingerrui,wangjingdong\}}@baidu.com \\
}

\begin{document}
\maketitle

\begin{abstract}
DETR accomplishes end-to-end object detection 
through iteratively generating multiple object candidates based on image features and
promoting one candidate for each ground-truth object.
The traditional training procedure using one-to-one supervision in the original DETR lacks direct supervision for the object detection candidates.

We aim at
improving the DETR training efficiency
by explicitly supervising the candidate generation procedure
through mixing one-to-one supervision
and one-to-many supervision.
Our approach, namely MS-DETR, 
is simple,
and places one-to-many supervision
to the object queries
of the primary decoder
that is used for inference.
In comparison to existing DETR variants with one-to-many supervision, such as Group DETR and
Hybrid DETR,
our approach does not need additional decoder branches or object queries; 
the object queries of the primary decoder in our approach directly benefit
from one-to-many supervision
and thus
are superior in object candidate prediction.
Experimental results show
that our approach outperforms 
related DETR variants, such as DN-DETR,
Hybrid DETR, and Group DETR,
and the combination with related DETR variants further improves the performance.
Code will be available at: \url{https://github.com/Atten4Vis/MS-DETR}.

\end{abstract}

\section{Introduction}

Detection Transformer (DETR)~\cite{carion2020end},
an end-to-end object detection approach,
has been attracting a lot of research attention~\cite{li2022exploring,misra2021end,liu2023petrv2,liu2022petr,yang2023bevformer}.
It is composed of 
a CNN backbone, a transformer encoder,
and a transformer decoder.
The decoder
is a stack of decoder layers,
and each layer consists of self-attention,
cross-attention and FFNs,
followed by class and box predictor.

\begin{figure}[t]
    \centering
    \footnotesize
    \begin{subfigure}{0.325\linewidth}
        \includegraphics[width=\linewidth]{fig-1/fig_1_1.png}
    \end{subfigure}
    \begin{subfigure}{0.325\linewidth}
        \includegraphics[width=\linewidth]{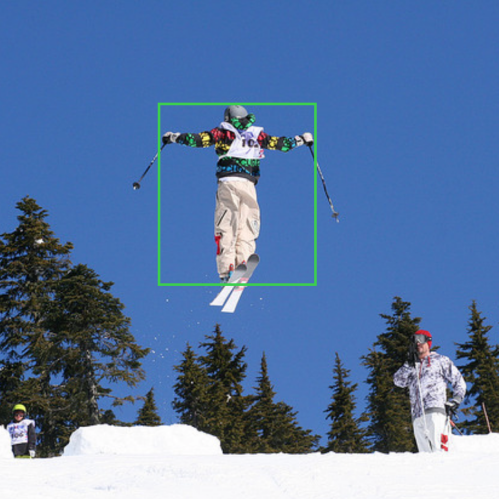}
    \end{subfigure}
    \begin{subfigure}{0.325\linewidth}
        \includegraphics[width=\linewidth]{fig-1/fig_1_3.png}
    \end{subfigure}
    
    \vspace{2pt}
    
    \begin{subfigure}{0.325\linewidth}
        \includegraphics[width=\linewidth]{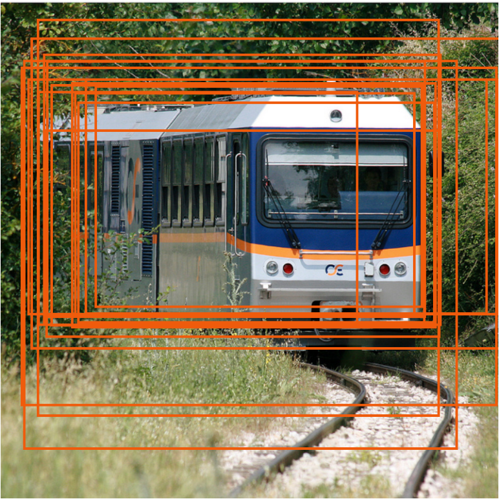}
    \end{subfigure}
    \begin{subfigure}{0.325\linewidth}
        \includegraphics[width=\linewidth]{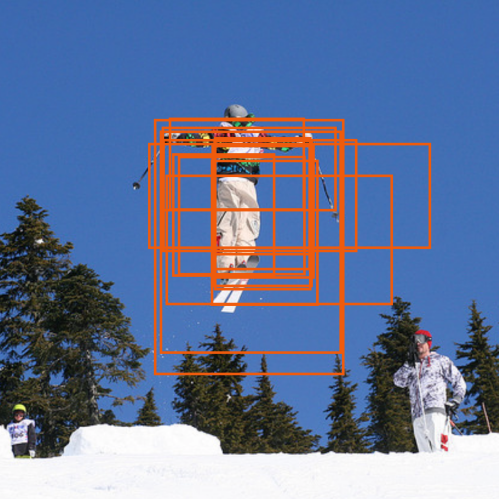}
    \end{subfigure}
    \begin{subfigure}{0.325\linewidth}
        \includegraphics[width=\linewidth]{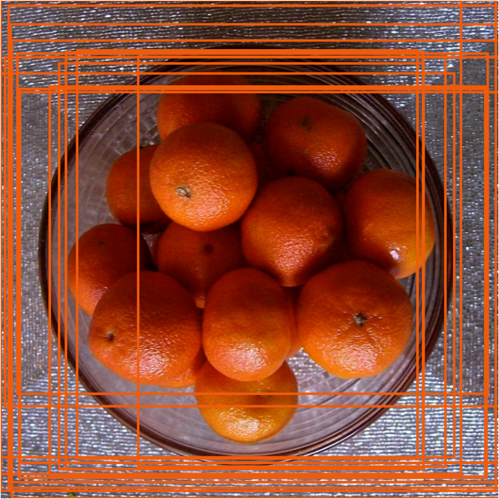}
    \end{subfigure}
    
    \vspace{2pt}
    
    \begin{subfigure}{0.325\linewidth}
        \includegraphics[width=\linewidth]{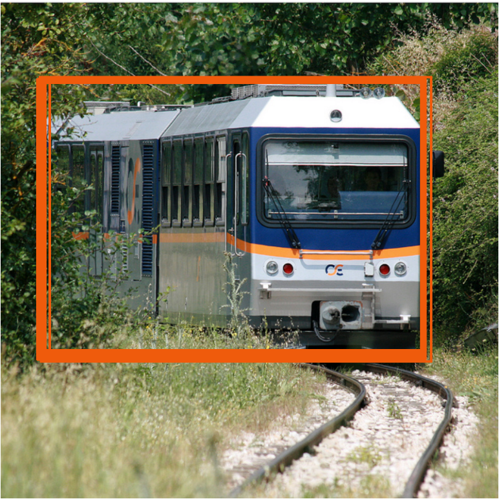}
    \end{subfigure}
    \begin{subfigure}{0.325\linewidth}
        \includegraphics[width=\linewidth]{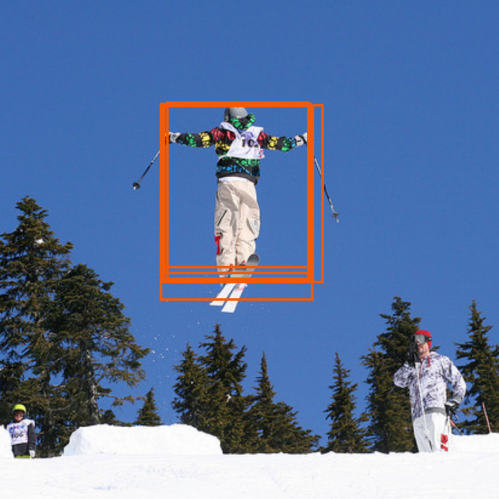}
    \end{subfigure}
    \begin{subfigure}{0.325\linewidth}
        \includegraphics[width=\linewidth]{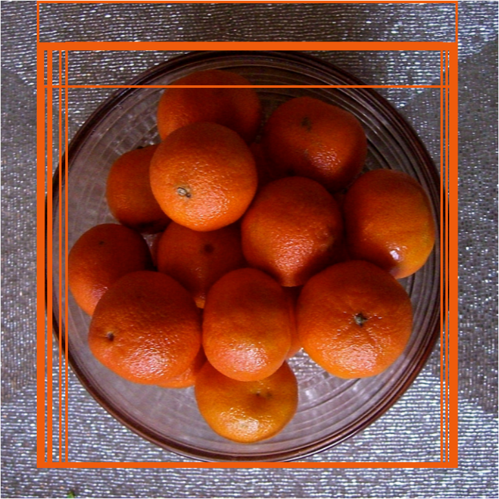}
    \end{subfigure}
    
    \caption{\textbf{Mixed supervision
    leads to better detection candidates.}
    {Top}: ground-truth box. 
    {Middle}: candidate boxes
    from top-$20$ queries with the baseline. 
    {Bottom}: candidate boxes from top-$20$ queries with our MS-DETR.
    One can see that MS-DETR generates better detection candidates than the baseline.}
    \label{fig:effect}
    \vspace{-3mm}
\end{figure}

The DETR decoder generates multiple object candidates
that are represented 
in the form of object queries,
and promotes one candidate and demotes other duplicate candidates for each ground-truth object 
in an end-to-end learning manner~\cite{carion2020end}.
The duplicate candidates, which are close to the ground-truth object,
are illustrated in Figure~\ref{fig:effect}.
The role of candidate generation is mainly taken by decoder cross-attention.
The role of candidate de-duplication is mainly taken by decoder self-attention
together with one-to-one supervision,
ensuring the selection of a single candidate for each ground-truth object.
Unlike NMS-based methods (e.g., Faster R-CNN~\cite{faster-rcnn})
which usually introduce a supervision
for candidate generation,
the DETR training procedure lacks explicit supervision for generating multiple object detection candidates.

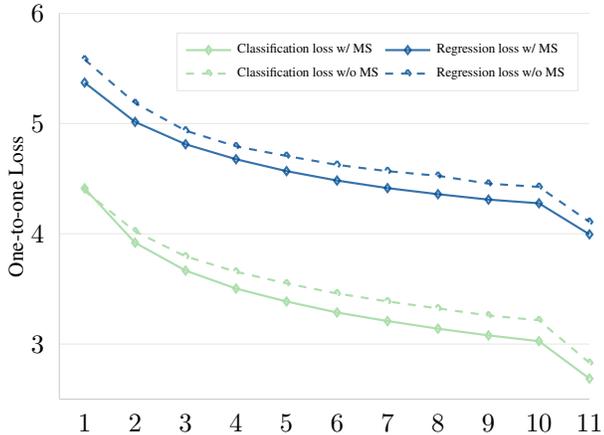
\begin{figure}[t]
    \begin{tikzpicture}
    \begin{axis}[
    legend columns=2, 
    legend style={at={(0.60,0.95)},anchor=north, font=\tiny, draw=gray!20}, legend cell align={left},
    y label style={at={(-0.05,0.5)}},
    ylabel={One-to-one Loss}, ymajorgrids=true, xtick={0, 1,2,3,4,5,6,7,8,9,10,11}, xticklabels={$0$, $1$, $2$, $3$, $4$, $5$, $6$, $7$, $8$, $9$, $10$, $11$},
    ytick={2,3,4,5,6,7,8,9,10,11,12},
    tick style={draw=none},
    y label style={font=\footnotesize},
    height=8cm,
    width=9cm,
    y post scale=0.8,
    x post scale=0.95,
    axis lines = left,
    every outer y axis line/.style={draw=gray!40},
    every outer x axis line/.style={draw=gray!40},
    grid style={line width=.1pt, draw=gray!20},
    xmin=0.5, ymax=6, ymin=2.5]

        \addplot[line width=0.8pt, mark size=1.5pt, mark=diamond, draw=mossgreen, draw opacity=1.0] coordinates {
            (1,4.411924850536783)(2,3.918472108741937)(3,3.666231836547681)(4,3.502698312185621)(5,3.3855006473221314)(6,3.2857318614915276)(7,3.208150114058384)(8,3.139122548385755)(9,3.0782398354026794)(10,3.025955504994907)(11,2.6866085289937307)
        };

        \addplot[line width=0.8pt, mark size=1.5pt, mark=diamond, draw=mediumelectricblue, draw opacity=0.8] coordinates {
            (1,5.3706677630432855)(2,5.014165653228503)(3,4.811792883096343)(4,4.67570943611712)(5,4.567931372550881)(6,4.482541141658921)(7,4.41387925714851)(8,4.358828202070386)(9,4.309652052642358)(10,4.276200302351349)(11,3.994427511119583)
        };

        \addplot[line width=0.8pt, mark size=1.5pt, mark=diamond, draw=mossgreen, dashed, draw opacity=1.0] coordinates {
            (1,4.390336978525008)
            (2,4.021612338029401)
            (3,3.7947109630856706)
            (4,3.655459507930977)
            (5,3.549697775629043)
            (6,3.4581078194231947)
            (7,3.386419574603746)
            (8,3.3229440690159198)
            (9,3.258638192151116)
            (10,3.2149863719867624)
            (11,2.8292147293979126)
        };

        \addplot[line width=0.8pt, mark size=1.5pt, mark=diamond, draw=mediumelectricblue, dashed, draw opacity=0.8] coordinates {
            (1,5.580625278724315)(2,5.185410410081133)(3,4.9349262084407295)(4,4.7924564241139525)(5,4.705276953324274)(6,4.624698260352348)(7,4.5674979818565875)(8,4.525387909093561)(9,4.454360629313613)(10,4.4247232134521095)(11,4.107419457317334)
        };
        \addlegendentry{Classification loss w/ MS}
        \addlegendentry{Regression loss w/ MS}
        \addlegendentry{Classification loss  w/o MS}
        \addlegendentry{Regression loss w/o MS}
        \end{axis}
    \end{tikzpicture}
    \caption{\textbf{Mixed supervision leads to lower one-to-one losses than the baseline}.
  The $x$-axis corresponds to \#epochs, and the $y$-axis corresponds to the training loss from one-to-one supervision.
  Dashed and solid lines correspond to the loss curve of the Deformable DETR baseline and the MS-DETR, respectively. Best viewed in color.
  }
  \label{fig:11losscomparison}
  \vspace{-5mm}
\end{figure}

\begin{figure*}[t]
  \centering
  \footnotesize
(a)~~\includegraphics[scale=0.85]{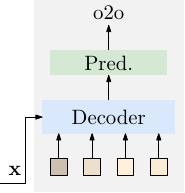}~~
(b)~~\includegraphics[scale=0.85]{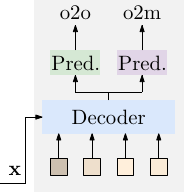}~~
(c)~~\includegraphics[scale=0.85]{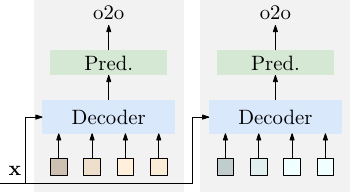}~~
(d)~~\includegraphics[scale=0.85]{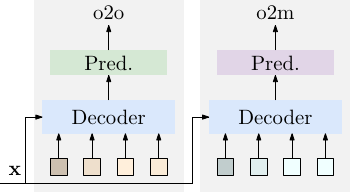}
   \caption{
   \textbf{Illustrating the architecture
   differences.}
   (a) Original DETR. It is trained with one-to-one supervision.
   (b) Our MS-DETR.
   It is trained by mixing one-to-one
   and one-to-many supervision. 
   The two supervisions
   are both imposed on the primary decoder.
   (c) Group DETR and DN-DETR.
   Additional parallel decoders are introduced, and 
   one-to-one supervision is imposed
   on the additional decoders.
   More additional decoders are possibly used in Group DETR and DN-DETR.
   (d) Hybrid DETR. 
   An additional parallel decoder is 
   added,
   and one-to-many supervision is imposed
   on the additional decoder.}
  \label{fig:mixed111m}
  \vspace{-3mm}
\end{figure*}

We propose to supervise the queries of the primary decoder with the mixture of one-to-one supervision and additional one-to-many supervision to improve the training efficiency.
The architecture is very simple.
We add a module similar to the prediction head for one-to-one supervision,
that consists of one box predictor,
one class predictor
for one-to-many supervision.
The resulting approach,
named MS-DETR,
is illustrated in
Figure~\ref{fig:mixed111m}.
We want to point out that
the additional modules only influence the training process
and the inference process remains unchanged.

Figure~\ref{fig:effect} illustrates 
how the extra supervision
influences the candidate detection.
We observe that the DETR without one-to-many supervision also generates multiple candidates for each ground-truth object.
After using the additional one-to-many supervision,
one can see that
the predicted boxes are better,
implying that the candidates are better.
We observe that
the one-to-one classification and box regression losses
are decreased
when the additional one-to-many supervision is added (as illustrated in Figure~\ref{fig:11losscomparison}).
This provides evidence that
one-to-many supervision
is able to improve the candidates
and thus is helpful 
for optimizing the one-to-one supervision loss.

Our approach improves the quality of object queries by introducing additional supervision for collecting information from image features.
It is distinct from and complements related training-efficient schemes~\cite{deformable-detr,dab,conditional,conditionalv2,yao2021efficient,sun2021rethinking}, such as conditional DETR~\cite{conditional} and Deformable DETR~\cite{deformable-detr}, which modify cross-attention architectures or change the query forms.
Our approach is related to, yet clearly different from DETR variants employing one-to-many supervision~\cite{DN,Group,Hybrid}. 
Specifically, our approach directly imposes one-to-many supervision on queries of the primary decoder.
In contrast, Group DETR and Hybrid DETR apply supervision to queries in additional decoders other than the primary decoder.
The differences from closely-related methods
are illustrated in Figure~\ref{fig:mixed111m}.

Experimental results show that
our approach achieves consistent improvements
over DETR-based methods, including DETR variants with modified cross-attention or query formulation (Deformable DETR~\cite{deformable-detr}, DAB-DETR~\cite{dab}),
as well as other training-efficient variants (DN-DETR~\cite{DN},
Group DETR~\cite{Group}, Hybrid-DETR~\cite{Hybrid}).
Combining our approach
with other DETR variants with one-to-many supervision, 
such as Group DETR and Hybrid DETR,
is able to further improve the performance,
indicating that our approach is complementary to these variants.
In addition,
it is observed that our 
approach is more computation and memory-efficient as our approach does not include additional decoder branches and object queries.

\section{Related Work}
\noindent\textbf{Decoder cross-attention 
and query formulation modification.}
Cross-attention performs interactions
between image features and current object queries
to refine detection candidates
that are represented 
in the form of object queries.

Deformable DETR~\cite{deformable-detr} uses deformable attention, an extension of deformable convolution~\cite{dai2017deformable},
that selects the highly informative regions,
to replace the original cross-attention architecture.
Conditional DETR~\cite{conditional} separates
the spatial and content queries
and computes the
spatial attention to softly select the informative regions.
SMCA~\cite{gao2021fast} uses the Gaussian-like weight for spatial attention computation.
DAB-DETR~\cite{dab} and Conditional DETR v2~\cite{conditionalv2} use boxes to represent the position of queries.
Anchor DETR~\cite{anchordetr} uses anchor boxes to serve as the predefined reference region to aid in detecting objects of varying scales.

\vspace{1mm}
\noindent\textbf{One-to-many supervision
with parallel decoders.}
One-to-many supervision assigns
one ground-truth object to 
multiple object queries
to speed up DETR training.
Existing methods depend on
the additional parallel weight-sharing decoder.

DN-DETR~\cite{DN} introduces parallel weight-shared decoders
with each decoder handling a group of 
of noisy queries that are formed by
by adding noises to 
ground-truth objects\footnote{Initially DN-DETR
is motivated by one-to-one assignment stabilization.
We discuss it from another perspective:
parallel decoders and one-to-many supervision.}.
Group DETR~\cite{Group,chen2022group} instead learns the object queries of the additional decoders.
DN-DETR and Group DETR perform one-to-one supervision for each group of object queries,
resulting in one-to-many supervision 
for all the groups of object queries.
DINO~\cite{DINO} has a similar idea to DN-DETR, where contrastive denoising queries are introduced for group-wise one-to-one supervision.
DQS~\cite{zhang2023dense} adds a parallel dense query branch with one-to-many supervision alongside its distinct query branch.
Hybrid DETR~\cite{Hybrid} adds one additional parallel decoder
with additional object queries,
where one-to-many supervision is directly conducted
on the additional decoder.

Our approach MS-DETR is related to those methods
in that MS-DETR also introduces one-to-many supervision. 
MS-DETR clearly differs
from those methods
that do not modify the supervision
for the original (primary) decoder.
MS-DETR does not introduce additional decoders,
additional queries,
and performs one-to-many supervision merely 
on the original decoder.

DETA~\cite{DETA} directly performs one-to-many supervision
on the same decoder with extra decoders and queries.
Unfortunately, it removes one-to-one supervision
and brings NMS back as post-processing. 
The mixture of one-to-one and one-to-many supervisions
on the single decoder
without using NMS
is underexplored.

\vspace{1mm}
\noindent\textbf{One-to-many supervision in traditional methods.}
One-to-many assignment is widely adopted 
in deep learning approach for object detection~\cite{he2018mask,focal,zhang2020bridging,ge2021yolox,wang2023yolov7,redmon2016you,yolo,li2021generalized,li2020generalized,zhou2019objects}.
For example,
Faster R-CNN~\cite{faster-rcnn} 
and FCOS~\cite{tian2019fcos}
forms the objective function
by assigning multiple anchors and multiple center pixels
for one ground-truth,
followed by NMS postprocessing~\cite{neubeck2006efficient}
for duplicate removal.

Our approach is partially
inspired by the resemblance
between DETR and traditional methods~\cite{wang2021scaled,bochkovskiy2020yolov4,sun2021sparse,zheng2020distance}:
the DETR decoder finds the candidates
through cross-attention,
interacting
with image features,
and filter out duplicate candidates
through self-attention and one-to-one supervision.
The latter part is similar to NMS postprocessing,
and the former part is like most detectors.
Thus, we introduce one-to-many supervision
to the DETR decoder
for improving the candidate quality.

\begin{figure*}[t]
  \centering
  \footnotesize
(a)~~\includegraphics[scale=1]{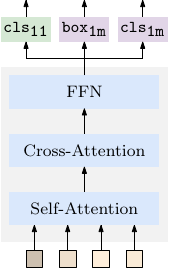}~~~
(b)~~\includegraphics[scale=1]{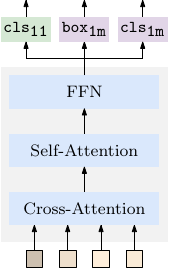}~~~
(c)~~\includegraphics[scale=1]{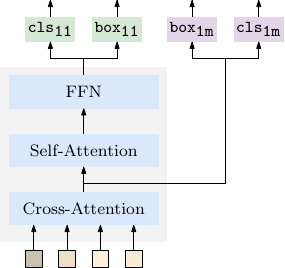}~~~
(d)~~\includegraphics[scale=1]{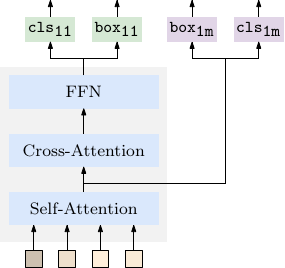}
   \caption{\textbf{MS-DETR implementations}. 
   (a) One-to-one and one-to-many supervisions are conducted
   on the output object queries for each decoder layer.
   (b) The two supervisions are conducted
   on the output object queries for each decoder layer that is slightly modified: first perform cross-attention and then self-attention.
   (c) and (d) The one-to-many supervisions are conducted 
   on the internal object queries.
   $\texttt{cls}_{\texttt{11}}$ and
   $\texttt{box}_{\texttt{11}}$
   are class and box predictors for
   one-to-one supervision,
   and
   $\texttt{cls}_{\texttt{1m}}$,
   and $\texttt{box}_{\texttt{1m}}$
   are class and box predictors
   for one-to-many supervision.
   The image features input to cross-attention
   are not depicted for clarity.}  
   \label{fig:MSDETRImplementations}
   \vspace{-3mm}
\end{figure*}

\section{MS-DETR}
\subsection{Preliminaries}
\noindent\textbf{DETR architecture.}
The initial DETR architecture consists of a CNN and transformer encoder,
a transformer decoder,
and object class and box position predictors.

The input image $\mathbf{I}$
goes through the encoder,
getting the image features: 
$$\mathbf{X} = \texttt{Encoder}(\mathbf{I}).$$
The learnable object queries 
$\mathbf{Q}$ and the image features $\mathbf{X}$
are fed into the decoder, 
resulting in the final object queries: 
$$\tilde{\mathbf{Q}} = \texttt{Decoder}(\mathbf{Q}).$$
The object queries are parsed
to the boxes and the classification 
scores\footnote{
The classification score 
is a combination of
the degree that the candidate belongs to
one class
and the degree that
it is better than other (duplicate)
candidates.}
through the predictors:
\begin{align}
\mathbf{B}
=~\texttt{box}_{\texttt{11}}(\tilde{\mathbf{Q}}),\quad \mathbf{S} =~\texttt{cls}_{\texttt{11}}(\tilde{\mathbf{Q}}).
\end{align}
For brevity, we use the subscript $\texttt{11}$ and $\texttt{1m}$ to indicate one-to-one and one-to-many respectively.

\vspace{1mm}
\noindent\textbf{Decoder.}
The transformer decoder
is a stack of decoder layers.
There are two main layers:
a self-attention layer,
which collects the information of other queries (candidates) for each query for duplicate candidate removal,
a cross-attention layer,
which collects the object candidates from image features 
in the form of queries
that are fed into an FFN layer
and then the box and class predictors.

\vspace{1mm}
\noindent\textbf{One-to-one supervision.}
The original DETR is trained 
with the one-to-one supervision.
One candidate prediction corresponds to 
one ground-truth object, and vice versa,
\begin{align}(\mathbf{y}_{\sigma(1)}, \bar{\mathbf{y}}_1), 
(\mathbf{y}_{\sigma(2)}, \bar{\mathbf{y}}_2), 
,
\dots,
(\mathbf{y}_{\sigma(N)}, \bar{\mathbf{y}}_N),
\end{align}
where $\sigma(\cdot)$
is the optimal permutation of $N$ indices,
and $[\bar{\mathbf{y}}_1~\bar{\mathbf{y}}_2, \dots~
\bar{\mathbf{y}}_N] = \bar{\mathbf{Y}}$ 
corresponds to ground truth,
and $\mathbf{y} = [\mathbf{s}^\top~ \mathbf{b}^\top]^\top$.

The one-to-one loss function is written as follows:
\begin{align}
    \mathcal{L}_{11} 
    = \sum\nolimits_{n=1}^N
    (\ell_{c11}(s_{\sigma(n)}, \bar{s}_n)
    + \ell_{b11}(\mathbf{b}_{\sigma(n)}, \bar{\mathbf{b}}_n)),
\end{align}
where $\ell_{c11}(\cdot)$
is the classification loss,
and $\ell_{b11}(\cdot)$
is the box regression loss.

The one-to-one supervision helps suppress duplicate candidates and promote a single candidate per ground-truth object,
by collecting information from other candidates through self-attention and comparing each candidate to the collected information.
One-to-one supervision and 
self-attention performing interactions
between object queries
jointly take the role of NMS
typically employed in traditional object detection methods.

\subsection{Mixed Supervision}

\noindent\textbf{One-to-many supervision.}
One-to-many supervision
is used
in traditional detection methods
to learn and provide better candidates
for the NMS post-processing.
For example,
Faster R-CNN dynamically assigns the ground-truth object to predicted boxes if they have enough overlap with the ground-truth object.
FCOS assigns the ground-truth object
to the pixels at the object center.

In light of the resemblance 
between NMS 
and
the joint role
of self-attention and one-to-one supervision, %
we propose to use one-to-many assignment supervision
to explicitly improve the quality of object queries
and accordingly the detection candidates. %
We adopt an additional module
for one-to-many prediction:
\begin{align}
\mathbf{B}
=~ \texttt{box}_{\texttt{1m}}(\tilde{\mathbf{Q}}),\quad\mathbf{S} =~ \texttt{cls}_{\texttt{1m}}(\tilde{\mathbf{Q}})
\end{align}

The one-to-many loss function is given as:
\begin{align}
    \mathcal{L}_{1m} 
    = &\sum\nolimits_{n=1}^N
    \sum\nolimits_{i=1}^{K_n}(\ell_{c1m}(s_{n_i}, \bar{s}_n)
    + \ell_{b1m}(\mathbf{b}_{n_i}, \bar{\mathbf{b}}_n)),\nonumber
\end{align}
where $\{(s_{n_1},\mathbf{b}_{n_1}), 
(s_{n_2}, \mathbf{b}_{n_2}), 
\dots, (s_{n_{K_n}}, \mathbf{b}_{n_{K_n}})\}$
are assigned to the $n$th ground-truth object. $K_n$ is the number of the matched predictions for the $n$-th ground-truth object.

\vspace{1mm}
\noindent\textbf{One-to-many matching.}
\label{sec:o2m_matching}
One-to-many matching is based on
the matching score between
a prediction $(\mathbf{s}, \mathbf{b})$
(from the one-to-many predictors)
and the ground-truth
$(\bar{c}, \bar{\mathbf{b}})$,
which is a combination 
of IoU and classification scores:
\begin{align}
\texttt{MatchScore}
(\mathbf{s}, \mathbf{b}, \bar{c}, \bar{\mathbf{b}})= 
\alpha s_{\bar{c}}
+
(1-\alpha)\texttt{IoU}(\mathbf{b}, \bar{\mathbf{b}}),\nonumber
\end{align}

Following~\cite{DETA}, we select the top $K$ queries
in terms of the matching scores 
for each ground-truth object,
and then filter out the queries
if their matching scores are lower than
a threshold $\tau$,
forming the matching query set. 
We also include the query obtained
from one-to-one matching 
into the matching query set for each ground-truth, which brings a slight-better gain ($+0.2$ mAP).

\subsection{Implementation}
\label{sec:impl}
The additional module for one-to-many supervision consists of box and class predictors, which are identical to those used in one-to-one supervision. %
The box predictor is implemented as a three-layer MLP with ReLU activation, 
and the class predictor is implemented as a single linear layer.

A straightforward implementation 
(Figure~\ref{fig:MSDETRImplementations} (a)) is 
to perform one-to-many prediction
over the output object queries of each decoder layer,
which is similar to one-to-one prediction.
We merge the one-to-one box prediction
into the one-to-many box prediction.
The loss function 
for one ground-truth object 
consists of
three parts:
one-to-one classification loss,
one-to-many box regression loss,
and one-to-many classification loss:
\begin{align}
\ell_{c11}(s_{\sigma(n)}, \bar{s}_n) 
    + \sum\nolimits_{i=1}^{K_n}(\ell_{c1m}(s_{n_i}, \bar{s}_n)
    + \ell_{b1m}(\mathbf{b}_{n_i}, \bar{\mathbf{b}}_n))\nonumber
\end{align}

Considering that the role of DETR cross-attention
is to generate multiple candidates 
according to image features
and the role of self-attention is to collect the information
of other candidates for duplicate removal,
we change the order
of the components in the decoder layer
from \texttt{self-attention} $\rightarrow$ \texttt{cross-attention} $\rightarrow$ \texttt{FFN}
to \texttt{cross-attention} $\rightarrow$ \texttt{self-attention} $\rightarrow$ \texttt{FFN}.
This (illustrated in Figure~\ref{fig:MSDETRImplementations} (b)) is similar to traditional methods, such as Faster R-CNN:
first generate multiple candidates for each object
and then remove duplicate candidates
using NMS.
This almost does not influence the performance\footnote{The change makes a large influence
if there are fewer decoder layers.}.

\begin{table*}[t]
  \centering
    \setlength{\tabcolsep}{4.5pt}
    \renewcommand{\arraystretch}{1.25}
    \footnotesize
  \caption{\textbf{Comparison of MS-DETR against other methods with one-to-many (O2M) supervision on various baselines.} MS-DETR consistently improves various popular DETR baselines. Compared with other O2M methods, our improvement is comparable (usually larger). \textcolor{gray}{Baseline} denotes the results of the baselines without any O2M methods applied. $^*$ denotes using auxiliary denoising queries, where the number of queries is a rough approximation.}
  \begin{tabular}{l|l|c|cc|c|cccccc}
    \shline
    Baseline & O2M method & \#epochs & \makecell[c]{\#queries \\ (primary)} & \makecell[c]{\#queries \\ (extra)} & \makecell[c]{extra \\ decoder branch} & mAP & AP$_{50}$ & AP$_{75}$ & AP$_s$ & AP$_m$ & AP$_l$ \\\hline
    \multirow{4}{*}{DAB-Deformable-DETR} & \textcolor{gray}{Baseline} & \textcolor{gray}{12} & \textcolor{gray}{$300$} & \textcolor{gray}{0} & \textcolor{gray}{\xmark} & \textcolor{gray}{$44.2$} & \textcolor{gray}{62.5} & \textcolor{gray}{47.3} & \textcolor{gray}{$27.5$} & \textcolor{gray}{$47.1$} & \textcolor{gray}{$58.6$}  \\
        & Group-DETR  & 12 & 300 & 3000 & \cmark & $45.7$ & - & - & $28.1$ & $49.0$ & $60.6$ \\
        & DN-DETR & 12 & $300$ & $70^*$ & \cmark & $46.0$ & $63.8$ & $49.9$ & $27.7$ & $49.1$ & $62.3$  \\
        & MS-DETR & 12  & $300$ & 0 & \xmark & $\mathbf{47.9~\textcolor{mycolor}{(+3.7)}}$ & 
$65.1$ & $51.6$ & $30.1$ & $51.2$ & $63.2$  \\\hline
    
    \multirow{3}{*}{Deformable DETR} & \textcolor{gray}{Baseline}   & \textcolor{gray}{12} & \textcolor{gray}{$300$} & \textcolor{gray}{0} & \textcolor{gray}{\xmark} &  \textcolor{gray}{$43.7$} & \textcolor{gray}{$62.2$} & \textcolor{gray}{$46.9$} & \textcolor{gray}{$26.4$} & \textcolor{gray}{$46.4$} & \textcolor{gray}{$57.9$} \\
        & H-DETR & 12 & 300 & 1500 & \cmark & $45.9$ & - & - & - & - & -  \\
        & MS-DETR & 12 & $300$ & 0 & \xmark & $\mathbf{47.6~\textcolor{mycolor}{(+3.7)}}$ & $64.9$ & $51.7$ & $29.6$ & $50.9$ & $63.3$ \\\hline
    
    \multirow{3}{*}{Deformable DETR++} & \textcolor{gray}{Baseline} & \textcolor{gray}{12}  & \textcolor{gray}{$300$} & \textcolor{gray}{0} & \textcolor{gray}{\xmark} & \textcolor{gray}{$47.0$} & \textcolor{gray}{$65.3$} &\textcolor{gray}{$51.0$} & \textcolor{gray}{$30.1$} & \textcolor{gray}{$50.5$} & \textcolor{gray}{$60.7$}  \\
        & H-DETR & 12 & 300 & 1500 & \cmark &  $48.7$ & $66.4$ & $52.9$ & $31.2$ & $51.5$ & $63.5$  \\
        & MS-DETR  & 12 & 300 & 0 & \xmark & $\mathbf{48.8~\textcolor{mycolor}{(+1.8)}}$ & $66.2$ & $53.2$ & $31.5$ & $52.3$ & $63.7$  \\\hline

    \multirow{3}{*}{Deformable DETR++} & \textcolor{gray}{Baseline} & \textcolor{gray}{12} & \textcolor{gray}{900} & \textcolor{gray}{0} & \textcolor{gray}{\xmark} &  \textcolor{gray}{$47.6$} & \textcolor{gray}{$65.8$} & \textcolor{gray}{$51.8$} & \textcolor{gray}{$31.2$} & \textcolor{gray}{$50.6$} & \textcolor{gray}{$62.6$} \\
        & DINO  & 12 & $900$ & $200^*$ & \cmark & $49.0$ & 
$66.6$ & $53.5$ & $32.0$ & $52.3$ & $63.0$  \\
        & MS-DETR & 12 & $900$ & 0 & \xmark &  $\mathbf{50.0~\textcolor{mycolor}{(+2.4)}}$ & $67.3$ & $54.4$ & $31.6$ & $53.2$ & $64.0$ \\\hline

    \multirow{3}{*}{Deformable DETR++} & \textcolor{gray}{Baseline}  & \textcolor{gray}{24} & \textcolor{gray}{$900$} & \textcolor{gray}{0} & \textcolor{gray}{\xmark} &  \textcolor{gray}{$49.8$}  & \textcolor{gray}{$67.0$} & \textcolor{gray}{$54.2$} & \textcolor{gray}{$31.4$} & \textcolor{gray}{$52.8$} & \textcolor{gray}{$64.1$} \\
        & DINO  & 24 & $900$ & $200^*$ & \cmark & $50.4$ & 
$68.3$ & $54.8$ & $33.3$ & $53.7$ & $64.8$  \\
        & MS-DETR & 24 & $900$ & 0 & \xmark &  $\mathbf{50.9~\textcolor{mycolor}{(+1.1)}}$ & $68.4$ & $56.1$ & $34.7$ & $54.3$ & $65.1$ \\
    \shline
  \end{tabular}
  \label{tab:comparisontoo2m}
  \vspace{-3mm}
\end{table*}

We then place the one-to-many supervision
over the internal object queries
processed with an FFN output from cross-attention (illustrated in Figure~\ref{fig:MSDETRImplementations} (c)).
We assume that the internal object queries
within the decoder layer (after cross-attention)
contain much information about each individual candidate,
and the output object query 
of the decoder layer (after self-attention)
additionally contains the information 
about other candidates.
Thus, imposing one-to-many supervision over 
internal object queries (output from cross-attention) 
potentially benefits the training,
empirically verified in Table~\ref{tab:o2msupplacement}.

In contrast, placing one-to-many supervision
over internal object queries 
without exchanging the order
of cross-attention and self-attention
(in Figure~\ref{fig:MSDETRImplementations} (d))
leads to worse performance. 
The reason
might be that 
the supervision placement is not consistent
to the roles of cross-attention and self-attention:
cross-attention is mainly about generating multiple candidates, and self-attention
collects the information of other candidates
mainly for promoting the winning candidate.

\section{Experiments}
\subsection{Object Detection}
\noindent\textbf{Setting.}
We verify our approach on various representative DETR-based detectors,
such as DAB-DETR~\cite{dab}, 
Deformable DETR~\cite{deformable-detr} and its strong extension Deformable-DETR++~\cite{DINO, Hybrid} that is implemented with three additional tricks: mixed query selection, look forward twice, and zero dropout rate.
We report the results in comparison
to and in combination
with representative DETR variants
with one-to-many supervision, including 
DN-DETR~\cite{DN},
Hybrid DETR~\cite{Hybrid},
Group DETR~\cite{Group}, and DINO~\cite{DINO}. 
We use ResNet-50~\cite{resnet} as the CNN backbone.
The models are trained 
mainly for $12$ epochs
and partially for $24$ epochs.
The models are trained on the COCO {\texttt {train2017}} and evaluated on the COCO {\texttt {val2017}}.
Implementation details are in the Supplemental Material.

\vspace{1mm}
\noindent\textbf{Comparison against DETR variants with one-to-many supervision.}
The results are reported 
in Table~\ref{tab:comparisontoo2m}.
MS-DETR brings consistent improvements on different DETR baselines.
Specifically, the gains over DAB-Deformable-DETR, Deformable DETR, and Deformable DETR++ are $3.7$, $3.7$, and $1.8$ in terms of mAP under 12 epochs. 

In comparison to DETR variants
with one-to-many supervision,
the gains of our approach 
are greater than Group DETR and
DN-DETR based on DAB-Deformable-DETR: $+1.5$ mAP vs $+3.7$ AP, $+1.8$ mAP vs $+3.7$ mAP.
The gains are also greater 
than Hybrid DETR based on Deformable DETR 
and
Deformable DETR++: $+2.2$ mAP vs $+3.7$ mAP, $+1.7$ mAP vs $+1.8$ mAP.
In comparison to DINO,
a strong method with one-to-many supervision
with denoising queries, our improvement is also greater: $+1.4$ mAP vs $+2.4$ mAP and $+0.8$ mAP vs $+1.1$ mAP, for 12 epochs and 24 epochs, respectively.

The superiority over DETR variants
with one-to-many supervision
comes from that
that our approach
impose one-to-many supervision
directly to the object queries in the primary decoder.

\begin{table*}[t]
  \centering
    \setlength{\tabcolsep}{14pt}
    \renewcommand{\arraystretch}{1.25}
    \footnotesize
  \caption{\textbf{Combination with other methods with one-to-many supervision.} MS-DETR is a complementary approach to existing O2M methods and consistently improves performance.}
  \begin{tabular}{l|c|c|cccccc}
    \shline
    Model & w/ MS-DETR & \#epochs & mAP & AP$_{50}$ & AP$_{75}$ & AP$_s$ & AP$_m$ & AP$_l$ \\
    \hline
    DN-DETR &  & $12$ & $41.7$ & $61.4$ & $44.1$ & $21.2$ & $45.0$ & $60.2$     \\
    DN-DETR & \cmark & $12$ & $43.7~\textcolor{mycolor}{(+\mathbf{2.0})}$ & $62.5$ & $46.6$ & $22.3$& $47.9$ & $62.0$  \\
    \hline
    DAC-DETR &  & $12$ & $47.1$ & $64.8$ & $51.1$ & $29.2$ & $50.6$ & $62.4$  \\
    DAC-DETR & \cmark  & $12$ & $48.1~\textcolor{mycolor}{(+\mathbf{1.0})}$ & $65.6$ & $52.3$ & $30.5$ & $51.2$ & $63.0$     \\
    \hline
    Group-DETR &  & $12$ & $48.0$ & $66.4$ & $52.2$ & $30.9$ & $51.1$ & $63.2$  \\
    Group-DETR & \cmark  & $12$ & $48.6~\textcolor{mycolor}{(+\mathbf{0.6})}$ & $66.0$ & $53.1$ & $30.3$ & $52.2$ & $64.1$     \\
    \hline
    {H}-DETR &  & $12$  & $48.7$ & $66.4$ & $52.9$ & $31.2$ & $51.5$ & $63.5$     \\
    {H}-DETR & \cmark  & $12$ & $49.5~\textcolor{mycolor}{(+\mathbf{0.8})}$ & $67.0$ & $53.8$ & $31.2$ & $52.7$ & $64.0$     \\
    \hline
    DINO &  & $12$ & $49.0$ & $66.6$ & $53.5$ & $32.0$ & $52.3$ & $63.0$     \\
    DINO & \cmark  & $12$ & $50.3~\textcolor{mycolor}{(+\mathbf{1.3})}$ & $67.4$ & $55.1$ &  $32.7$ & $54.0$ & $64.6$     \\
    \hline
    DINO &  & $24$ & $50.4$ & $68.3$ & $54.8$ & $33.3$ & $53.7$ & $64.8$     \\
    DINO & \cmark  & $24$ & $51.7~\textcolor{mycolor}{(+\mathbf{1.3})}$ & $68.7$ & $56.5$ & $34.0$ & $55.4$ & $65.5$     \\
    \shline
  \end{tabular}
  \label{tab:comb_o2m}
\end{table*}

\vspace{1mm}
\noindent\textbf{Combination with DETR variants with one-to-many supervision.}
Table \ref{tab:comb_o2m} shows the results combining our MS-DETR with other methods with one-to-many supervision. Our method consistently improves the performance of these methods. It brings $2.0$, $0.6$, $1.0$, $0.8$ and $1.3$ gains in mAP over DN-DETR(-DC5)~\cite{DN}, Group-DETR~\cite{Group}, DAC-DETR~\cite{dac}, Hybrid DETR~\cite{Hybrid} and DINO~\cite{DINO} under 12 epochs schedule, respectively.
Our approach further improves the performance of DINO by $1.3$ mAP under a longer training schedule (24 epochs).

These methods apply one-to-many supervision on extra queries in the extra decoder branch(es), while the queries in the primary decoder branch are still supervised in a one-to-one manner.
Differently, our method directly applies one-to-many supervision on the queries in the primary decoder branch, thus achieving good complementary to these methods.

\begin{figure*}
\centering
\footnotesize
\begin{tikzpicture}[font=\footnotesize]
\pgfplotsset{compat=1.11,
    /pgfplots/ybar legend/.style={
    /pgfplots/legend image code/.code={%
       \draw[##1,/tikz/.cd,yshift=-0.3em]
        (0cm,0cm) rectangle (7pt,0.8em);},
   },
}
\begin{axis}[
    width=0.4\linewidth, 
    height=0.25\linewidth,
    ybar,
    symbolic x coords={4, 5, 6, 7, 8},
    xtick=data,
    ylabel=mAP,
    ymin=44.8,
    ymax=47.8,
    bar width=12pt,
    legend style={at={(0.02,0.96)},anchor=north west},
    nodes near coords,
    nodes near coords style={font=\tiny},
    axis y line=none,
    axis x line =bottom,
    axis line style={-},
    enlarge x limits=0.2,
    tick style={draw=none},
    y post scale=0.66,
    x post scale=0.72,
    legend style={at={(0.5, 1.05)}, anchor=north, legend columns =-1, draw=none, fill=none, font=\tiny},
    ]
    \addplot[bar shift=0pt,fill=mediumelectricblue, draw=mediumelectricblue, fill opacity=0.5, draw opacity=0.5] coordinates {(4, 47.2) (5, 47.4) (6, 47.6) (7, 47.3) (8, 47.2)};
    \end{axis}
\node[below left, yshift=12pt, font=\footnotesize] at (current bounding box.south west) {(a)~~};
\end{tikzpicture}
\quad
\hspace{40pt}
\begin{tikzpicture}[font=\footnotesize]
\pgfplotsset{compat=1.11,
    /pgfplots/ybar legend/.style={
    /pgfplots/legend image code/.code={%
       \draw[##1,/tikz/.cd,yshift=-0.3em]
        (0cm,0cm) rectangle (7pt,0.8em);},
   },
}
\begin{axis}[
    width=0.4\linewidth, 
    height=0.25\linewidth,
    ybar,
    symbolic x coords={0.2, 0.3, 0.4, 0.5, 0.6},
    xtick=data,
    ylabel=mAP,
    ymin=44.8,
    ymax=47.8,
    bar width=12pt,
    legend style={at={(0.02,0.96)},anchor=north west},
    nodes near coords,
    nodes near coords style={font=\tiny},
    axis y line=none,
    axis x line =bottom,
    axis line style={-},
    enlarge x limits=0.2,
    tick style={draw=none},
    y post scale=0.66,
    x post scale=0.72,
    legend style={at={(0.5, 1.05)}, anchor=north, legend columns =-1, draw=none, fill=none, font=\tiny},
    ]
    \addplot[bar shift=0pt,fill=mossgreen2, draw=mossgreen2, fill opacity=0.5, draw opacity=0.5] coordinates {(0.2, 47.3) (0.3, 47.5) (0.4, 47.6) (0.5, 47.2) (0.6, 46.9)};
    \end{axis}
\node[below left, yshift=12pt, font=\footnotesize] at (current bounding box.south west) {(b)~~};
\end{tikzpicture}
\quad
\hspace{40pt}
\begin{tikzpicture}[font=\footnotesize]
\pgfplotsset{compat=1.11,
    /pgfplots/ybar legend/.style={
    /pgfplots/legend image code/.code={%
       \draw[##1,/tikz/.cd,yshift=-0.3em]
        (0cm,0cm) rectangle (7pt,0.8em);},
   },
}
\begin{axis}[
    width=0.4\linewidth, 
    height=0.25\linewidth,
    ybar,
    symbolic x coords={0.1, 0.2, 0.3, 0.4, 0.5},
    xtick=data,
    ylabel=mAP,
    ymin=44.8,
    ymax=47.8,
    bar width=12pt,
    legend style={at={(0.02,0.96)},anchor=north west},
    nodes near coords,
    nodes near coords style={font=\tiny},
    axis y line=none,
    axis x line =bottom,
    axis line style={-},
    enlarge x limits=0.2,
    tick style={draw=none},
    y post scale=0.66,
    x post scale=0.72,
    legend style={at={(0.5, 1.05)}, anchor=north, legend columns =-1, draw=none, fill=none, font=\tiny},
    ]
    \addplot[bar shift=0pt,fill=lightsalmon, draw=lightsalmon, fill opacity=0.5, draw opacity=0.5] coordinates {(0.1, 47.1) (0.2, 47.4) (0.3, 47.6) (0.4, 47.3) (0.5, 47.0)};
    \end{axis}
\node[below left, yshift=12pt, font=\footnotesize] at (current bounding box.south west) {(c)~~};
\end{tikzpicture}
\caption{\textbf{Influence of the hyper-parameters for one-to-many assignment}. (a) Influence of $K$ for selecting top-$K$ positive queries, (b) Influence of the threshold $\tau$ to filter out low-quality queries, and (c) Influence of the matching score weight $\alpha$.}
\label{fig:hyper_o2m_assign}
\vspace{-5mm}
\end{figure*}
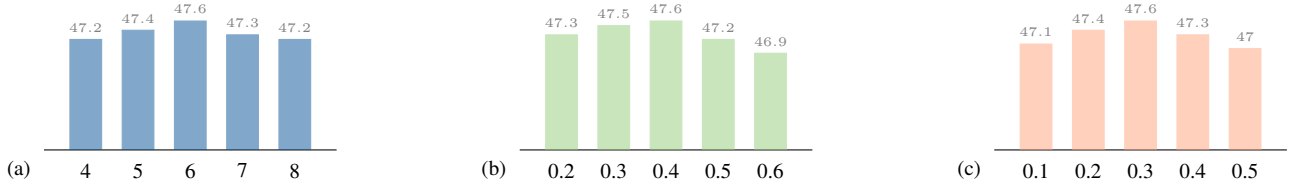

\begin{table}[t]
  \centering
    \setlength{\tabcolsep}{4pt}
    \renewcommand{\arraystretch}{1.25}
    \footnotesize
  \caption{\textbf{Comparison of training time cost and memory cost.} 
  The costs of DETR variants with one-to-many supervision, Hybrid DETR and Group DETR are reported. 
  The baseline method is Deformable-DETR++ with ResNet50 backbone. 
  All the methods are trained with the same batch size and on the same machine with $8 \times$ V100 GPUs. Training time is the average training time of one epoch. 
  }

  \begin{tabular}{l|cc|cccc}
    \shline
    Method & \makecell[c]{\#queries \\ (primary)} & \makecell[c]{\#queries \\ (extra)} & training time & GPU Memory \\
    \hline
     Baseline & 300 & 0  & $67$min & $5116$M \\
     Hybrid DETR & 300 & 1500  & $103$min & $8680$M \\
     Group DETR & 300 & 1500  & $95$min & $7128$M \\
     MS-DETR & 300 & 0 & $69$min & $5243$M \\
    \shline
  \end{tabular}
  \label{tab:computationmemoryefficiency}
  \vspace{-3mm}
\end{table}

\vspace{1mm}
\noindent\textbf{Computation and memory efficiency.}
Table~\ref{tab:computationmemoryefficiency}
reports the computation cost and the memory cost
of the baseline (Deformable DETR++ with $300$ queries),
Hybrid DETR,
Group DETR,
and our MS-DETR.
The batch sizes are the same for all the methods.
The training time of each epoch 
is obtained by averaging the time over $12$ epochs.

One can see that for our approach MS-DETR, 
the additional time 
from the one-to-many supervision
is minor: increase $2$ minutes
from the time cost of the baseline.
In contrast, the additional time costs of Group DETR and Hybrid DETR are $+36$ and $+28$ minutes, much larger than our approach.

Our approach is also more memory-efficient.
For instance, our method incurs only a minor increase of 127M memory ($\sim$2\%) relative to the baseline, while Hybrid DETR and Group DETR lead to substantial memory increases of nearly 60\% and 40\% respectively, in relation to the baseline. 
The reason is that Hybrid DETR and Group DETR introduce more queries and thus more computation overhead.

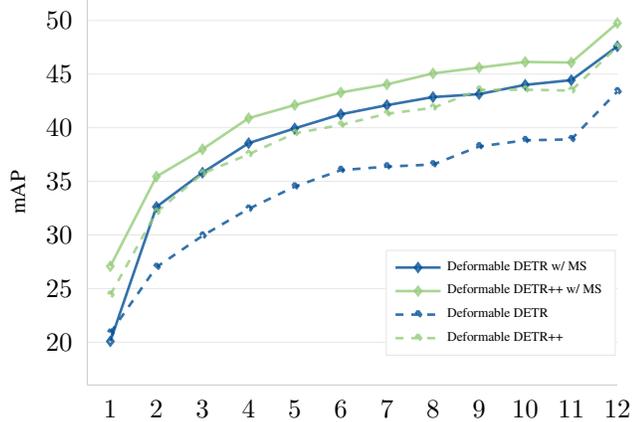
\begin{figure}[t]
  \centering
\begin{tikzpicture}
    \begin{axis}[
    legend columns=1, 
    legend style={at={(0.78,0.35)},anchor=north, font=\tiny, draw=gray!20}, legend cell align={left},
    y label style={at={(-0.10,0.5)}},
    ylabel={mAP}, ymajorgrids=true, xtick={ 1,2,3,4,5,6,7,8,9,10,11, 12}, xticklabels={$1$, $2$, $3$, $4$, $5$, $6$, $7$, $8$, $9$, $10$, $11$, $12$},
    ytick={20, 25, 30, 35, 40, 45, 50},
    tick style={draw=none},
    y label style={font=\footnotesize},
    height=8cm,
    width=9cm,
    y post scale=0.8,
    x post scale=0.95,
    axis lines = left,
    every outer y axis line/.style={draw=gray!40},
    every outer x axis line/.style={draw=gray!40},
    grid style={line width=.1pt, draw=gray!20},
    xmin=0.5, ymax=52, ymin=16]
        \addplot[line width=1.0pt, mark size=1.5pt, mark=diamond, draw=mediumelectricblue, draw opacity=0.8] coordinates {
            (1, 20.078250485302676)
            (2, 32.616847741690647)
            (3, 35.809170238674476)
            (4, 38.56156024318435)
            (5, 39.940714685252726)
            (6, 41.253915528996743)
            (7, 42.10430466509844)
            (8, 42.861725237828113)
            (9, 43.13678487538484)
            (10, 44.002555181504294)
            (11, 44.43860190805197)
            (12, 47.59456593441559)
        };
        
        \addplot[line width=1.0pt, mark size=1.5pt, mark=diamond, draw=mossgreen2, draw opacity=0.8] coordinates {
            (1, 27.07817402756708)
            (2, 35.445487941386)
            (3, 37.99142330356418)
            (4, 40.8934286940801)
            (5, 42.11609645405951)
            (6, 43.291780634877947)
            (7, 44.04582442257059)
            (8, 45.0694170218479)
            (9, 45.61496597962808)
            (10, 46.136520833272426)
            (11, 46.084799714613783)
            (12, 49.76483276985912)
        };

        \addplot[line width=1.0pt, mark size=1.5pt, mark=diamond, dashed, draw=mediumelectricblue, draw opacity=0.8] coordinates {
            (1, 20.867181728686932)
            (2, 26.99935348187379)
            (3, 29.92314700339679)
            (4, 32.43175568572885)
            (5, 34.504860911105245)
            (6, 36.0422057323059)
            (7, 36.37546118356249)
            (8, 36.568617989730856)
            (9, 38.2528891459586)
            (10, 38.828013036306447)
            (11, 38.91632146193518)
            (12, 43.382580699933265)
        };

        \addplot[line width=1.0pt, mark size=1.5pt, mark=diamond, dashed, draw=mossgreen2, draw opacity=0.8] coordinates {
            (1, 24.417628915529854)
            (2, 32.120583130669733)
            (3, 35.72367891933596)
            (4, 37.524347526238744)
            (5, 39.476986735066416)
            (6, 40.26071302010572)
            (7, 41.305355715233866)
            (8, 41.85794626597533)
            (9, 43.530525836866496)
            (10, 43.548925616909967)
            (11, 43.46884983470889)
            (12, 47.55118929740858)
        };

        \addlegendentry{Deformable DETR w/ MS}
        \addlegendentry{Deformable DETR++ w/ MS}
        \addlegendentry{Deformable DETR}
        \addlegendentry{Deformable DETR++}

        \end{axis}
    \end{tikzpicture}
  \caption{\textbf{Convergence curves.}
  MS-DETR accelerates the training process for DETR variants. Dashed and solid lines correspond to the baseline method and our MS counterparts. 
  The $x$-axis corresponds to \#epochs and the $y$-axis corresponds to the mAP evaluated on COCO \texttt{val2017}. }
  \label{fig:convergencecurves}
  \vspace{-3mm}
\end{figure}

\vspace{1mm}
\noindent\textbf{Convergence curves.}
In Figure~\ref{fig:convergencecurves}, 
we present the convergence curve of our MS-DETR alongside 
its corresponding baselines, 
the Deformable DETR with $300$ queries and Deformable DETR++ with $900$ queries. The models utilize ResNet-50 as the backbone architecture and undergo 12 epochs of training.
We observe that the introduction of mixed supervision in our MS-DETR accelerates the training convergence.

\noindent\textbf{Combination with IoU-aware loss.} 
We study the combination of MS-DETR with another line of work improving DETR with IoU-aware loss~\cite{align,liu2023detection,pu2023rank,wu2022iou,zhang2021varifocalnet}. 
We apply our approach over Align-DETR~\cite{align} based on DINO baseline.
Table~\ref{tab:align} shows that MS-DETR improves the performance of Align-DETR by $0.5$ AP and {$0.6$} AP under 12 and 24 epochs training schedules respectively.
This shows that MS-DETR is also complementary to IoU-aware loss.

\begin{table}[t]
  \centering
    \setlength{\tabcolsep}{2.5pt}
    \renewcommand{\arraystretch}{1.25}
    \footnotesize
  \caption{\textbf{Combination of MS-DETR with Align-DETR.} 
  Our approach further improves the Align-DETR performance.
  The Align-DETR includes the one-to-many loss like Hybrid DETR. }
    \begin{tabular}{c|c|c|c|c} 
    \shline
    \#epochs & w/ Align Loss & w/ Hybrid Loss & w/ MS-DETR & mAP \\
    \hline
     12 & \cmark & \cmark & & $50.2$ \\
     12 & \cmark & & \cmark & $50.7~\textcolor{mycolor}{(+\mathbf{0.5})}$  \\
    \hline
     24 & \cmark & \cmark & &$51.3$ \\
     24 & \cmark & & \cmark & $51.9~\textcolor{mycolor}{(+\mathbf{0.6})}$  \\
    \shline
    \end{tabular}
  \label{tab:align}
  \vspace{-5mm}
\end{table}

\subsection{Ablation Study}
\noindent\textbf{Hyperparameters in one-to-many matching.}
We illustrate the influence of the three hyperparameters in one-to-many matching. 

Figure~\ref{fig:hyper_o2m_assign} (a) illustrates the impact of the hyperparameter $K$ on the selection of top-$K$ queries. We empirically find that our approach achieves optimal performance when $K=6$. A small value of $K$ decreases the number of positive queries. A large value of $K$ causes the object imbalance problem~\cite{DETA,oksuz2020imbalance}.

Figure~\ref{fig:hyper_o2m_assign} (b) visualizes the influence of the threshold $\tau$ 
that is used to filter out low-quality queries for one-to-many supervision. 
Our approach achieves its best results when $\tau=0.4$. 
Lowering the value of $\tau$ increases the inclusion of low-quality queries, while raising it reduces the number of positive queries eligible for one-to-many supervision.

In Figure~\ref{fig:hyper_o2m_assign} (c), we present the impact of the score weight $\alpha$ in the one-to-many match score. 
A higher value of $\alpha$ will increase the importance of the classification score and a lower value of $\alpha$ will increase the importance of the IoU score.
We empirically find our method achieves the best performance when $\alpha$ is set to $0.4$.

\begin{table}[t]
  \centering
    \setlength{\tabcolsep}{5.5pt}
    \renewcommand{\arraystretch}{1.25}
    \footnotesize
    \caption{\textbf{The effect
    of one-to-many supervision placement.}
    Deformable DETR is used as the baseline.
    The four ways for placing one-to-many supervision
    are illustrated in Figure~\ref{fig:MSDETRImplementations}.
    Output of \texttt{FFN} = the output object queries of each decoder layer.
    Internal output = the internal object queries
    output from the first attention of each decoder layer.}
    \begin{tabular}{l|c|c|c}
        \shline
         & decoder configuration & queries for supervision & mAP \\
        \shline
        Baseline & $-$ & $-$ & $43.7$ \\
        \hline
        (a) & \texttt{SA} 
        $\rightarrow$ \texttt{CA} $\rightarrow$ \texttt{FFN}  & Output of \texttt{FFN} & $47.0$ \\
        (b) & \texttt{CA} $\rightarrow$ \texttt{SA} $\rightarrow$ \texttt{FFN}& Output of \texttt{FFN} & $47.1$ \\
        (c) & \texttt{CA} $\rightarrow$ \texttt{SA} $\rightarrow$ \texttt{FFN}& Internal output 
        of \texttt{CA} & $\mathbf{47.6}$ \\
        (d) & \texttt{SA} $\rightarrow$ \texttt{CA} $\rightarrow$ \texttt{FFN} & Internal output of \texttt{SA}& $46.1$ \\
        \shline
    \end{tabular}
    \label{tab:o2msupplacement}
    \vspace{-3mm}
\end{table}

\vspace{1mm}
\noindent\textbf{One-to-many supervision placement.}
We report the empirical results 
for placing one-to-many supervision over
internal and output object queries 
in the decoder layer,
as well as the two order configurations of cross-attention
and self-attention in the layer.
The four variants are illustrated in
Figure~\ref{fig:MSDETRImplementations}.

The results are given in Table~\ref{tab:o2msupplacement}.
The four MS-DETR variants achieve
large gains
over the baseline. 
The simple variant,
directly placing the one-to-many supervision
over the output object queries of each decoder layer,
gets a gain of $3.3$ mAP,
and exchanging the order
of cross-attention and self-attention
does not influence the result.
If placing the one-to-many supervision
on the internal object queries output from cross-attention
for the configuration of $\texttt{cross-attention}
\rightarrow \texttt{self-attention}
\rightarrow \texttt{FFN}$,
a further gain $0.6$ is obtained.
This confirms the analysis in {Section~\ref{sec:impl}}.

\begin{figure*}[t]
  \centering
\includegraphics[width=1\textwidth]{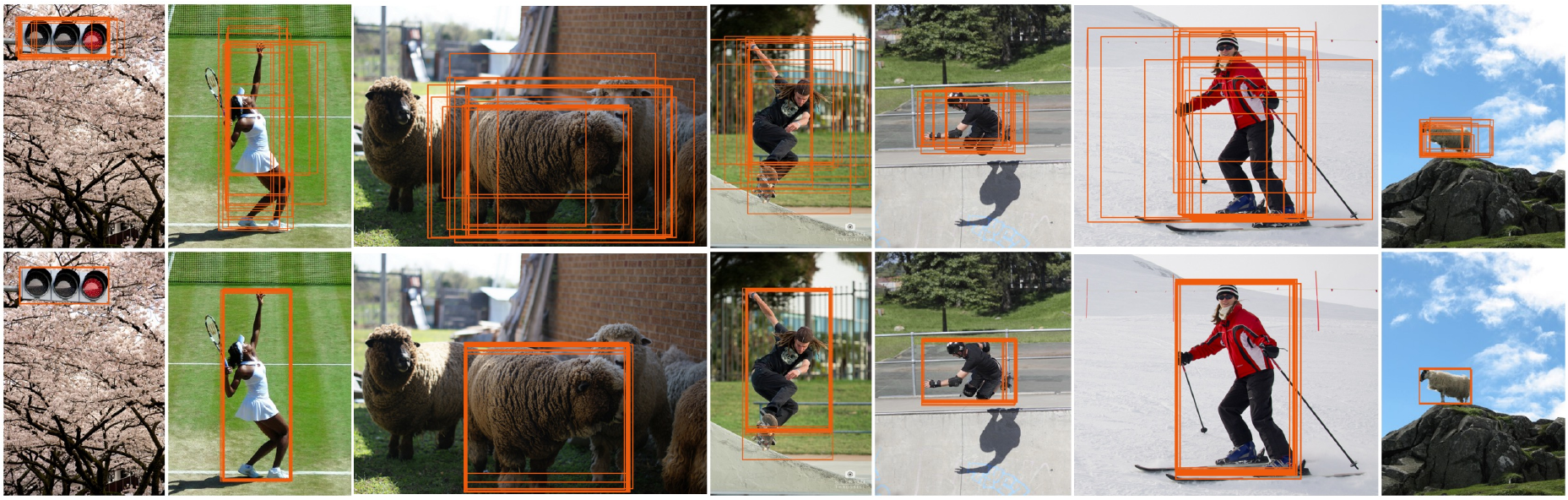}
  \caption{\textbf{More Examples illustrating better candidates from one-to-many supervision.} 
  We visualized the detection results of the top-$20$ candidates for one ground-truth object. Top: the Deformable DETR++ baseline trained with one-to-one supervision. 
  Bottom: our MS-DETR trained with mixed one-to-one and one-to-many supervision.
  One can see that the quality of the candidates is better with mixed supervision.}
  \label{fig:bettercandidates}
  \vspace{-5mm}
\end{figure*}

\vspace{1mm}
\noindent\textbf{Weight sharing
for predictors of one-to-many and one-to-one supervision.}
We perform empirical analysis 
for sharing weights of box and class predictors 
between one-to-many and one-to-one supervision.
The results are given in~Table~\ref{tab:weightsharing}.

One can see that sharing the weights
for both box and class predictors gets
the best performance.
It is relatively easily understandable 
for sharing the weights of the box predictor:
both the box predictors for one-to-one and one-to-many supervision
need to extract the same features
for box prediction,
and sharing weights in some sense
adds more supervision.

{
We assume that sharing the weights of the class predictor 
leads to 
(1) adding more supervision for training some weights in the predictor
that are useful for both
one-to-one and one-to-many classification,
(2) leaving the weights for scoring the duplicate candidates
learned from one-to-one supervision
which do not influence the prediction
for one-to-many supervision.}

\begin{table}[t]
  \centering
    \setlength{\tabcolsep}{13pt}
    \renewcommand{\arraystretch}{1.25}
    \footnotesize
  \caption{\textbf{The effect of sharing
  the weights} of box and class predictors between one-to-many supervision
  and one-to-one supervision.}
  \begin{tabular}{c|cccc}
    \shline
    box predictor & \xmark & \xmark & \cmark & \cmark \\
    
    cls predictor & \xmark & \cmark & \xmark &\cmark \\
    \hline
    mAP & $47.2$ & $47.0$ & ${47.4}$ & $\mathbf{47.6}$ \\
    \shline
  \end{tabular}
  \label{tab:weightsharing}
  \vspace{-2mm}
\end{table}

\vspace{1mm}
\noindent\textbf{Illustration
of better candidate prediction from one-to-many supervision.}
In Figure~\ref{fig:bettercandidates}, 
we present more examples to illustrate the improvement of candidate predictions achieved through one-to-many supervision. 
The predictions are obtained from the final object queries.
The detection results of the top-$20$ candidate queries 
with respect to the IoU scores are visualized.
In the top row, 
we showcase the detection results obtained by the Deformable DETR baseline, 
which is trained only with one-to-one supervision.
The bottom row displays detection results obtained by our MS-DETR.
One can see that the candidates produced under mixed supervision exhibit superior quality, demonstrating the effectiveness of our approach in enhancing the quality of the candidates.

\subsection{Application to Instance Segmentation}
We report the results
for the problem of instance segmentation, 
to further demonstrate the effectiveness.
We report the instance segmentation results over Mask-Deformable-DETR~\cite{Hybrid} baseline on the COCO-2017 \texttt{val} set.
We run experiments for $12$ and $50$ epochs based on ResNet50~\cite{resnet} backbone.
Table~\ref{tab:segmentation} shows that MS-DETR significantly improves the mask
mAP of the baseline by $3.2$ mAP under $12$ epochs training schedule.
It can still improve the mask mAP of the baseline by $2.5$ mAP under a longer $50$ epochs training schedule.

\section{Conclusion}
Our approach mixes an additional one-to-many supervision
with the original one-to-one supervision
for DETR training.
The improvement implies
that the additional one-to-many supervision
benefits the optimization for one-to-one supervision. 
One main characteristic is that
our approach explicitly supervises
the object queries.
Our approach is complementary to related methods
that mainly modify the cross-attention architecture
or learn the decoder weights
with additional queries or additional decoders.

\begin{table}[t]
  \centering
    \setlength{\tabcolsep}{10pt}
    \renewcommand{\arraystretch}{1.25}
    \footnotesize
  \caption{\textbf{Instance segmentation} on the COCO-2017 \textit{val} set \cite{ms-coco}. The results are obtained with ResNet50~\cite{resnet} and $12$ and $50$ epochs.}
  \vspace{-2mm}
    \begin{tabular}{c|c|c|cc} 
    \shline
    Epochs & w/ MS-DETR & Mask mAP & Box mAP \\
    \hline
    12 &  & 28.3 & 43.8 \\
    12 & \cmark & 31.5 \textcolor{mycolor}{$(+\mathbf{3.2})$} & 47.1 \textcolor{mycolor}{$(+\mathbf{3.3})$} \\
    \hline
    50  &  & 32.2 & 45.6 \\
    50 & \cmark & 34.7 \textcolor{mycolor}{$(+\mathbf{2.5})$} & 48.3 \textcolor{mycolor}{$(+\mathbf{2.7})$} \\
    \shline
    \end{tabular}
    \vspace{-2mm}
  \label{tab:segmentation}
\end{table}

{
    \small
    \bibliographystyle{ieeenat_fullname}
    \bibliography{main}
}

\end{document}

%% file: preamble.tex
\usepackage[dvipsnames]{xcolor}


%% file: main.bbl
\begin{thebibliography}{48}
\providecommand{\natexlab}[1]{#1}
\providecommand{\url}[1]{\texttt{#1}}
\expandafter\ifx\csname urlstyle\endcsname\relax
  \providecommand{\doi}[1]{doi: #1}\else
  \providecommand{\doi}{doi: \begingroup \urlstyle{rm}\Url}\fi

\bibitem[Bochkovskiy et~al.(2020)Bochkovskiy, Wang, and Liao]{bochkovskiy2020yolov4}
Alexey Bochkovskiy, Chien-Yao Wang, and Hong-Yuan~Mark Liao.
\newblock Yolov4: Optimal speed and accuracy of object detection.
\newblock \emph{arXiv preprint arXiv:2004.10934}, 2020.

\bibitem[Cai et~al.(2023)Cai, Liu, Wang, Ge, Zhang, and Huang]{align}
Zhi Cai, Songtao Liu, Guodong Wang, Zheng Ge, Xiangyu Zhang, and Di Huang.
\newblock Align-detr: Improving detr with simple iou-aware bce loss.
\newblock \emph{arXiv preprint arXiv:2304.07527}, 2023.

\bibitem[Carion et~al.(2020)Carion, Massa, Synnaeve, Usunier, Kirillov, and Zagoruyko]{carion2020end}
Nicolas Carion, Francisco Massa, Gabriel Synnaeve, Nicolas Usunier, Alexander Kirillov, and Sergey Zagoruyko.
\newblock End-to-end object detection with transformers.
\newblock In \emph{Computer Vision--ECCV 2020: 16th European Conference, Glasgow, UK, August 23--28, 2020, Proceedings, Part I 16}, pages 213--229. Springer, 2020.

\bibitem[Chen et~al.(2022{\natexlab{a}})Chen, Chen, Zeng, and Wang]{Group}
Qiang Chen, Xiaokang Chen, Gang Zeng, and Jingdong Wang.
\newblock Group detr: Fast training convergence with decoupled one-to-many label assignment.
\newblock \emph{arXiv preprint arXiv:2207.13085}, 2022{\natexlab{a}}.

\bibitem[Chen et~al.(2022{\natexlab{b}})Chen, Wang, Han, Zhang, Li, Chen, Chen, Wang, Han, Zhang, et~al.]{chen2022group}
Qiang Chen, Jian Wang, Chuchu Han, Shan Zhang, Zexian Li, Xiaokang Chen, Jiahui Chen, Xiaodi Wang, Shuming Han, Gang Zhang, et~al.
\newblock Group detr v2: Strong object detector with encoder-decoder pretraining.
\newblock \emph{arXiv preprint arXiv:2211.03594}, 2022{\natexlab{b}}.

\bibitem[Chen et~al.(2022{\natexlab{c}})Chen, Wei, Zeng, and Wang]{conditionalv2}
Xiaokang Chen, Fangyun Wei, Gang Zeng, and Jingdong Wang.
\newblock Conditional detr v2: Efficient detection transformer with box queries.
\newblock \emph{arXiv preprint arXiv:2207.08914}, 2022{\natexlab{c}}.

\bibitem[Dai et~al.(2017)Dai, Qi, Xiong, Li, Zhang, Hu, and Wei]{dai2017deformable}
Jifeng Dai, Haozhi Qi, Yuwen Xiong, Yi Li, Guodong Zhang, Han Hu, and Yichen Wei.
\newblock Deformable convolutional networks.
\newblock In \emph{Proceedings of the IEEE international conference on computer vision}, pages 764--773, 2017.

\bibitem[Gao et~al.(2021)Gao, Zheng, Wang, Dai, and Li]{gao2021fast}
Peng Gao, Minghang Zheng, Xiaogang Wang, Jifeng Dai, and Hongsheng Li.
\newblock Fast convergence of detr with spatially modulated co-attention.
\newblock In \emph{ICCV}, pages 3621--3630, 2021.

\bibitem[Ge et~al.(2021)Ge, Liu, Wang, Li, and Sun]{ge2021yolox}
Zheng Ge, Songtao Liu, Feng Wang, Zeming Li, and Jian Sun.
\newblock Yolox: Exceeding yolo series in 2021.
\newblock \emph{arXiv preprint arXiv:2107.08430}, 2021.

\bibitem[He et~al.(2016)He, Zhang, Ren, and Sun]{resnet}
Kaiming He, Xiangyu Zhang, Shaoqing Ren, and Jian Sun.
\newblock Deep residual learning for image recognition.
\newblock In \emph{Proceedings of the IEEE conference on computer vision and pattern recognition}, pages 770--778, 2016.

\bibitem[He et~al.(2018)He, Gkioxari, Dollár, and Girshick]{he2018mask}
Kaiming He, Georgia Gkioxari, Piotr Dollár, and Ross Girshick.
\newblock Mask r-cnn, 2018.

\bibitem[Hu et~al.(2023)Hu, Sun, Wang, and Yang]{dac}
Zhengdong Hu, Yifan Sun, Jingdong Wang, and Yi Yang.
\newblock Dac-detr: Divide the attention layers and conquer.
\newblock 2023.

\bibitem[Jia et~al.(2022)Jia, Yuan, He, Wu, Yu, Lin, Sun, Zhang, and Hu]{Hybrid}
Ding Jia, Yuhui Yuan, Haodi He, Xiaopei Wu, Haojun Yu, Weihong Lin, Lei Sun, Chao Zhang, and Han Hu.
\newblock Detrs with hybrid matching.
\newblock \emph{arXiv preprint arXiv:2207.13080}, 2022.

\bibitem[Li et~al.(2022{\natexlab{a}})Li, Zhang, Liu, Guo, Ni, and Zhang]{DN}
Feng Li, Hao Zhang, Shilong Liu, Jian Guo, Lionel~M Ni, and Lei Zhang.
\newblock Dn-detr: Accelerate detr training by introducing query denoising.
\newblock In \emph{Proceedings of the IEEE/CVF Conference on Computer Vision and Pattern Recognition}, pages 13619--13627, 2022{\natexlab{a}}.

\bibitem[Li et~al.(2020)Li, Wang, Wu, Chen, Hu, Li, Tang, and Yang]{li2020generalized}
Xiang Li, Wenhai Wang, Lijun Wu, Shuo Chen, Xiaolin Hu, Jun Li, Jinhui Tang, and Jian Yang.
\newblock Generalized focal loss: Learning qualified and distributed bounding boxes for dense object detection.
\newblock \emph{Advances in Neural Information Processing Systems}, 33:\penalty0 21002--21012, 2020.

\bibitem[Li et~al.(2021)Li, Wang, Hu, Li, Tang, and Yang]{li2021generalized}
Xiang Li, Wenhai Wang, Xiaolin Hu, Jun Li, Jinhui Tang, and Jian Yang.
\newblock Generalized focal loss v2: Learning reliable localization quality estimation for dense object detection.
\newblock In \emph{Proceedings of the IEEE/CVF Conference on Computer Vision and Pattern Recognition}, pages 11632--11641, 2021.

\bibitem[Li et~al.(2022{\natexlab{b}})Li, Mao, Girshick, and He]{li2022exploring}
Yanghao Li, Hanzi Mao, Ross Girshick, and Kaiming He.
\newblock Exploring plain vision transformer backbones for object detection.
\newblock In \emph{European Conference on Computer Vision}, pages 280--296. Springer, 2022{\natexlab{b}}.

\bibitem[Lin et~al.(2015)Lin, Maire, Belongie, Bourdev, Girshick, Hays, Perona, Ramanan, Zitnick, and Dollár]{ms-coco}
Tsung-Yi Lin, Michael Maire, Serge Belongie, Lubomir Bourdev, Ross Girshick, James Hays, Pietro Perona, Deva Ramanan, C.~Lawrence Zitnick, and Piotr Dollár.
\newblock Microsoft coco: Common objects in context, 2015.

\bibitem[Lin et~al.(2017)Lin, Goyal, Girshick, He, and Doll{\'a}r]{focal}
Tsung-Yi Lin, Priya Goyal, Ross Girshick, Kaiming He, and Piotr Doll{\'a}r.
\newblock Focal loss for dense object detection.
\newblock In \emph{Proceedings of the IEEE international conference on computer vision}, pages 2980--2988, 2017.

\bibitem[Liu et~al.(2022{\natexlab{a}})Liu, Li, Zhang, Yang, Qi, Su, Zhu, and Zhang]{dab}
Shilong Liu, Feng Li, Hao Zhang, Xiao Yang, Xianbiao Qi, Hang Su, Jun Zhu, and Lei Zhang.
\newblock Dab-detr: Dynamic anchor boxes are better queries for detr.
\newblock \emph{arXiv preprint arXiv:2201.12329}, 2022{\natexlab{a}}.

\bibitem[Liu et~al.(2023{\natexlab{a}})Liu, Ren, Chen, Zeng, Zhang, Li, Li, Huang, Su, Zhu, et~al.]{liu2023detection}
Shilong Liu, Tianhe Ren, Jiayu Chen, Zhaoyang Zeng, Hao Zhang, Feng Li, Hongyang Li, Jun Huang, Hang Su, Jun Zhu, et~al.
\newblock Detection transformer with stable matching.
\newblock \emph{arXiv preprint arXiv:2304.04742}, 2023{\natexlab{a}}.

\bibitem[Liu et~al.(2022{\natexlab{b}})Liu, Wang, Zhang, and Sun]{liu2022petr}
Yingfei Liu, Tiancai Wang, Xiangyu Zhang, and Jian Sun.
\newblock Petr: Position embedding transformation for multi-view 3d object detection.
\newblock In \emph{European Conference on Computer Vision}, pages 531--548. Springer, 2022{\natexlab{b}}.

\bibitem[Liu et~al.(2023{\natexlab{b}})Liu, Yan, Jia, Li, Gao, Wang, and Zhang]{liu2023petrv2}
Yingfei Liu, Junjie Yan, Fan Jia, Shuailin Li, Aqi Gao, Tiancai Wang, and Xiangyu Zhang.
\newblock Petrv2: A unified framework for 3d perception from multi-camera images.
\newblock In \emph{Proceedings of the IEEE/CVF International Conference on Computer Vision}, pages 3262--3272, 2023{\natexlab{b}}.

\bibitem[Meng et~al.(2021)Meng, Chen, Fan, Zeng, Li, Yuan, Sun, and Wang]{conditional}
Depu Meng, Xiaokang Chen, Zejia Fan, Gang Zeng, Houqiang Li, Yuhui Yuan, Lei Sun, and Jingdong Wang.
\newblock Conditional detr for fast training convergence.
\newblock In \emph{Proceedings of the IEEE/CVF International Conference on Computer Vision}, pages 3651--3660, 2021.

\bibitem[Misra et~al.(2021)Misra, Girdhar, and Joulin]{misra2021end}
Ishan Misra, Rohit Girdhar, and Armand Joulin.
\newblock An end-to-end transformer model for 3d object detection.
\newblock In \emph{Proceedings of the IEEE/CVF International Conference on Computer Vision}, pages 2906--2917, 2021.

\bibitem[Neubeck and Van~Gool(2006)]{neubeck2006efficient}
Alexander Neubeck and Luc Van~Gool.
\newblock Efficient non-maximum suppression.
\newblock In \emph{18th international conference on pattern recognition (ICPR'06)}, pages 850--855. IEEE, 2006.

\bibitem[Oksuz et~al.(2020)Oksuz, Cam, Kalkan, and Akbas]{oksuz2020imbalance}
Kemal Oksuz, Baris~Can Cam, Sinan Kalkan, and Emre Akbas.
\newblock Imbalance problems in object detection: A review.
\newblock \emph{IEEE transactions on pattern analysis and machine intelligence}, 43\penalty0 (10):\penalty0 3388--3415, 2020.

\bibitem[Ouyang-Zhang et~al.(2022)Ouyang-Zhang, Cho, Zhou, and Kr{\"a}henb{\"u}hl]{DETA}
Jeffrey Ouyang-Zhang, Jang~Hyun Cho, Xingyi Zhou, and Philipp Kr{\"a}henb{\"u}hl.
\newblock Nms strikes back.
\newblock \emph{arXiv preprint arXiv:2212.06137}, 2022.

\bibitem[Pu et~al.(2023)Pu, Liang, Hao, Yuan, Yang, Zhang, Hu, and Huang]{pu2023rank}
Yifan Pu, Weicong Liang, Yiduo Hao, Yuhui Yuan, Yukang Yang, Chao Zhang, Han Hu, and Gao Huang.
\newblock Rank-detr for high quality object detection.
\newblock \emph{arXiv preprint arXiv:2310.08854}, 2023.

\bibitem[Redmon et~al.(2016{\natexlab{a}})Redmon, Divvala, Girshick, and Farhadi]{redmon2016you}
Joseph Redmon, Santosh Divvala, Ross Girshick, and Ali Farhadi.
\newblock You only look once: Unified, real-time object detection.
\newblock In \emph{Proceedings of the IEEE conference on computer vision and pattern recognition}, pages 779--788, 2016{\natexlab{a}}.

\bibitem[Redmon et~al.(2016{\natexlab{b}})Redmon, Divvala, Girshick, and Farhadi]{yolo}
Joseph Redmon, Santosh Divvala, Ross Girshick, and Ali Farhadi.
\newblock You only look once: Unified, real-time object detection.
\newblock In \emph{Proceedings of the IEEE conference on computer vision and pattern recognition}, pages 779--788, 2016{\natexlab{b}}.

\bibitem[Ren et~al.(2015)Ren, He, Girshick, and Sun]{faster-rcnn}
Shaoqing Ren, Kaiming He, Ross Girshick, and Jian Sun.
\newblock Faster r-cnn: Towards real-time object detection with region proposal networks.
\newblock \emph{Advances in neural information processing systems}, 28, 2015.

\bibitem[Sun et~al.(2021{\natexlab{a}})Sun, Zhang, Jiang, Kong, Xu, Zhan, Tomizuka, Li, Yuan, Wang, et~al.]{sun2021sparse}
Peize Sun, Rufeng Zhang, Yi Jiang, Tao Kong, Chenfeng Xu, Wei Zhan, Masayoshi Tomizuka, Lei Li, Zehuan Yuan, Changhu Wang, et~al.
\newblock Sparse r-cnn: End-to-end object detection with learnable proposals.
\newblock In \emph{Proceedings of the IEEE/CVF conference on computer vision and pattern recognition}, pages 14454--14463, 2021{\natexlab{a}}.

\bibitem[Sun et~al.(2021{\natexlab{b}})Sun, Cao, Yang, and Kitani]{sun2021rethinking}
Zhiqing Sun, Shengcao Cao, Yiming Yang, and Kris~M Kitani.
\newblock Rethinking transformer-based set prediction for object detection.
\newblock In \emph{Proceedings of the IEEE/CVF international conference on computer vision}, pages 3611--3620, 2021{\natexlab{b}}.

\bibitem[Tian et~al.(2019)Tian, Shen, Chen, and He]{tian2019fcos}
Zhi Tian, Chunhua Shen, Hao Chen, and Tong He.
\newblock Fcos: Fully convolutional one-stage object detection, 2019.

\bibitem[Wang et~al.(2021{\natexlab{a}})Wang, Bochkovskiy, and Liao]{wang2021scaled}
Chien-Yao Wang, Alexey Bochkovskiy, and Hong-Yuan~Mark Liao.
\newblock Scaled-yolov4: Scaling cross stage partial network.
\newblock In \emph{Proceedings of the IEEE/cvf conference on computer vision and pattern recognition}, pages 13029--13038, 2021{\natexlab{a}}.

\bibitem[Wang et~al.(2023)Wang, Bochkovskiy, and Liao]{wang2023yolov7}
Chien-Yao Wang, Alexey Bochkovskiy, and Hong-Yuan~Mark Liao.
\newblock Yolov7: Trainable bag-of-freebies sets new state-of-the-art for real-time object detectors.
\newblock In \emph{Proceedings of the IEEE/CVF Conference on Computer Vision and Pattern Recognition}, pages 7464--7475, 2023.

\bibitem[Wang et~al.(2021{\natexlab{b}})Wang, Zhang, Yang, and Sun]{anchordetr}
Yingming Wang, Xiangyu Zhang, Tong Yang, and Jian Sun.
\newblock Anchor detr: Query design for transformer-based detector.
\newblock \emph{national conference on artificial intelligence}, 2021{\natexlab{b}}.

\bibitem[Wu et~al.(2022)Wu, Yang, Wang, and Li]{wu2022iou}
Shengkai Wu, Jinrong Yang, Xinggang Wang, and Xiaoping Li.
\newblock Iou-balanced loss functions for single-stage object detection.
\newblock \emph{Pattern Recognition Letters}, 156:\penalty0 96--103, 2022.

\bibitem[Yang et~al.(2023)Yang, Chen, Tian, Tao, Zhu, Zhang, Huang, Li, Qiao, Lu, et~al.]{yang2023bevformer}
Chenyu Yang, Yuntao Chen, Hao Tian, Chenxin Tao, Xizhou Zhu, Zhaoxiang Zhang, Gao Huang, Hongyang Li, Yu Qiao, Lewei Lu, et~al.
\newblock Bevformer v2: Adapting modern image backbones to bird's-eye-view recognition via perspective supervision.
\newblock In \emph{Proceedings of the IEEE/CVF Conference on Computer Vision and Pattern Recognition}, pages 17830--17839, 2023.

\bibitem[Yao et~al.(2021)Yao, Ai, Li, and Zhang]{yao2021efficient}
Zhuyu Yao, Jiangbo Ai, Boxun Li, and Chi Zhang.
\newblock Efficient detr: Improving end-to-end object detector with dense prior.
\newblock \emph{arXiv preprint arXiv:2104.01318}, 2021.

\bibitem[Zhang et~al.(2021)Zhang, Wang, Dayoub, and Sunderhauf]{zhang2021varifocalnet}
Haoyang Zhang, Ying Wang, Feras Dayoub, and Niko Sunderhauf.
\newblock Varifocalnet: An iou-aware dense object detector.
\newblock In \emph{Proceedings of the IEEE/CVF conference on computer vision and pattern recognition}, pages 8514--8523, 2021.

\bibitem[Zhang et~al.(2022)Zhang, Li, Liu, Zhang, Su, Zhu, Ni, and Shum]{DINO}
Hao Zhang, Feng Li, Shilong Liu, Lei Zhang, Hang Su, Jun Zhu, Lionel~M Ni, and Heung-Yeung Shum.
\newblock Dino: Detr with improved denoising anchor boxes for end-to-end object detection.
\newblock \emph{arXiv preprint arXiv:2203.03605}, 2022.

\bibitem[Zhang et~al.(2020)Zhang, Chi, Yao, Lei, and Li]{zhang2020bridging}
Shifeng Zhang, Cheng Chi, Yongqiang Yao, Zhen Lei, and Stan~Z Li.
\newblock Bridging the gap between anchor-based and anchor-free detection via adaptive training sample selection.
\newblock In \emph{Proceedings of the IEEE/CVF conference on computer vision and pattern recognition}, pages 9759--9768, 2020.

\bibitem[Zhang et~al.(2023)Zhang, Wang, Wang, Pang, Lyu, Zhang, Luo, and Chen]{zhang2023dense}
Shilong Zhang, Xinjiang Wang, Jiaqi Wang, Jiangmiao Pang, Chengqi Lyu, Wenwei Zhang, Ping Luo, and Kai Chen.
\newblock Dense distinct query for end-to-end object detection.
\newblock In \emph{Proceedings of the IEEE/CVF Conference on Computer Vision and Pattern Recognition}, pages 7329--7338, 2023.

\bibitem[Zheng et~al.(2020)Zheng, Wang, Liu, Li, Ye, and Ren]{zheng2020distance}
Zhaohui Zheng, Ping Wang, Wei Liu, Jinze Li, Rongguang Ye, and Dongwei Ren.
\newblock Distance-iou loss: Faster and better learning for bounding box regression.
\newblock In \emph{Proceedings of the AAAI conference on artificial intelligence}, pages 12993--13000, 2020.

\bibitem[Zhou et~al.(2019)Zhou, Wang, and Kr{\"a}henb{\"u}hl]{zhou2019objects}
Xingyi Zhou, Dequan Wang, and Philipp Kr{\"a}henb{\"u}hl.
\newblock Objects as points.
\newblock \emph{arXiv preprint arXiv:1904.07850}, 2019.

\bibitem[Zhu et~al.(2020)Zhu, Su, Lu, Li, Wang, and Dai]{deformable-detr}
Xizhou Zhu, Weijie Su, Lewei Lu, Bin Li, Xiaogang Wang, and Jifeng Dai.
\newblock Deformable detr: Deformable transformers for end-to-end object detection.
\newblock \emph{arXiv preprint arXiv:2010.04159}, 2020.

\end{thebibliography}
